\documentclass{article}

\usepackage[utf8]{inputenc} 
\usepackage[T1]{fontenc}    
\usepackage{hyperref}       
\usepackage{url}            
\usepackage{booktabs} 
\usepackage{amsfonts} 
\usepackage{nicefrac}   
\usepackage{microtype}

\usepackage{natbib}
\usepackage{amsmath} 
\usepackage{amsthm}
\usepackage{mathtools}
\usepackage{bbm}
\usepackage{latexsym}
\usepackage{dsfont}
\usepackage{pifont}

\usepackage{verbatim}
\usepackage{hyperref}

\usepackage{color} 
\usepackage{tikz}
\usetikzlibrary{arrows}
\usetikzlibrary{calc}
\usetikzlibrary{positioning}
\usepackage{array,multirow,booktabs}

\usepackage{placeins}

\setcitestyle{authoryear,open={(},close={)}}

\graphicspath{{./Figures/} {./Experiments/}}

\theoremstyle{definition}

\newcommand{\Wcal}{\mathcal{W}}
\newcommand{\Scal}{\mathcal{S}}
\newcommand{\Acal}{\mathcal{A}}

\definecolor{bl}{RGB}{20,20,150}

\newcommand{\MI}{\operatorname{MI}}

\newcommand{\pos}{x}
\newcommand{\RLoc}{R_{\text{Loc}}}
\newcommand{\RMI}{R_{\MI}}
\newcommand{\rLoct}{r_{\text{Loc,t}}}
\newcommand{\rMIt}{r_{\text{MI,t}}}

\usepackage[
letterpaper,
textheight=9in,
textwidth=5.5in,
top=1in
]{geometry}

\usepackage{parskip}

\title{Information Theoretically Aided Reinforcement Learning for Embodied Agents}
\date{ }

\usepackage{authblk}

\author[1]{Guido Mont\'ufar\thanks{montufar@mis.mpg.de}}
\author[1]{Keyan Ghazi-Zahedi\thanks{zahedi@mis.mpg.de}}
\author[1,2,3]{Nihat Ay\thanks{nay@mis.mpg.de}}
\affil[1]{\small Max Planck Institute for Mathematics in the Sciences, 04103 Leipzig, Germany}
\affil[2]{Faculty of Mathematics and Computer Science, Leipzig University, 04009 Leipzig, Germany}
\affil[3]{Santa Fe Institute, Santa Fe, NM 87501, USA}

\begin{document}

\maketitle

\thispagestyle{empty}
\begin{abstract}
\noindent Reinforcement learning for embodied agents is a challenging problem. The accumulated reward to be optimized is often a very rugged function, and gradient methods are impaired by many local optimizers. We demonstrate, in an experimental setting, that incorporating an intrinsic reward can smoothen the optimization landscape while preserving the global optimizers of interest. We show that policy gradient optimization for locomotion in a complex morphology is significantly improved when supplementing the extrinsic reward by an intrinsic reward defined in terms of the mutual information of time consecutive sensor readings. 
\newline
\noindent
\emph{Keywords: } 
intrinsic motivation, reinforcement learning, predictive information, embodied systems, policy gradient, POMDP, non-convex optimization
\end{abstract}

\section{Introduction}

Reinforcement learning is a powerful framework for learning to act when one does not have explicit examples of good action sequences, or one has an idea of what a good behavior could look like, but it is difficult to infer actions that could generate that behavior. 
In reinforcement learning one still requires a reward signal. 
In some cases it is clear how to choose an appropriate reward signal. 
For a gambler in a casino this may be the number of chips earned or lost in one round. 
In some other cases it may be less clear how to choose the reward signal, and naive choices may not be useful. 
For instance, if the reward is too sparse, it may take too long for the agent to discover good actions. 
More importantly, a perfectly good looking definition of the reward, maximizing which surely produces the desired behavior, may involve a hopelessly complicated optimization landscape. 

One possible resort lies in the idea that it may be easier to solve a difficult problem if we first, or concurrently, learn to solve a different problem. 
Curriculum learning~\citep{Bengio:2009:CL:1553374.1553380} and transfer learning~\citep{5288526} can be regarded as examples of this general strategy. 
Intrinsic motivation has been proposed as a task independent reward, which may serve to initialize or aid optimizing policies for a particular task~\citep[for an overview see][]{baldassarre2013intrinsically}. 
In Section~\ref{sec:relatedwork} we give a brief overview on approaches that have been considered. 

In this paper we study the predictive information of sensor readings as an intrinsic reward signal supplementing an extrinsic task objective in reinforcement learning of embodied agents. 
As hinted above, our motivation for using a combined reward is that, for complex morphologies, the extrinsic reward is often a very intricate function of the policy, with many local optimizers. 
The idea is that combining the two objectives may help overcome the local optimizers. 
A high predictive information requires that all degrees of freedom are active and move in a coordinated manner. Therefore, we expect that many coordinated embodied behaviors are accompanied by high values of the predictive information. 
Certainly this also implies that many maximizers of the predictive information may not 
perform well at a particular task. 
By using geometric mixtures we obtain a combined objective function that retains only the concurrent optimizers of both functions. 
In this paper the predictive information is used as a multiplicative supplement in gradient based reinforcement learning of a complex embodied morphology. 
To the best of our knowledge this is the first paper that pursues this.  
Our experiments show that incorporating a suitable amount of predictive information in the objective function allows for significant improvements of the learning performance. 
Furthermore, the implementation is very simple, generally applicable, and scalable. 

In Section~\ref{sec:PI} we discuss the predictive information of time consecutive sensor readings and the combination of intrinsic and extrinsic rewards. 
In Section~\ref{sec:relatedwork} we comment on related work regarding intrinsic motivation and predictive information. 
In Section~\ref{sec:experiments} we present our experiments on a complex embodied system, 
and in Section~\ref{sec:conclusions} our conclusions.

\section{Predictive information as an intrinsic reward}
\label{sec:PI}

In reinforcement learning~\citep{suttonbarto98} one considers a system with world states $\Wcal$ and an agent that makes observations from a set $\Scal$ and chooses actions from a set $\Acal$. 
One usually considers discrete time Markovian systems where the state $w_{t+1}$ at time $t+1$ is distributed according to a fixed but unknown conditional probability distribution $\alpha(\cdot | w_t, a_t)$ given the state $w_t$ and action $a_t$ at the previous time step. 
The observation, or sensor reading, $s_t$ at time $t$ is distributed according to a fixed but unknown conditional probability distribution $\beta(\cdot|w_t)$. 
The sequence of world states $w_1,\ldots, w_T$ (or rather its probability) depends on the actions taken by the agent. 
At each time step $t$ the agent receives a reward signal $r_t$ that typically is expressed as a function of $w_t$ and $a_t$. 
The goal of learning is that, based on the observations and rewards, the agent discovers a policy such that the resulting behavior maximizes an accumulated reward, for instance the average reward $R = \frac{1}{T}\sum_{t=1}^{T}r_t$ for some $T\in\mathbb{N}$. 

For a temporal sequence of random variables, 
the predictive information~\citep{Bialek:2001:PCL:1121086.1121087} is defined as the mutual information of the past and the future. 
It can be interpreted as the reduction on uncertainty about the future given the past. 
We will consider the mutual information of two time consecutive sensor readings, 
\begin{equation}
\MI(S;S') 
:= \sum_{s,s'} p(s,s') \log\frac{p(s,s')}{p(s)p(s')} 
= H(S') - H(S'|S). 
\end{equation}
This is the entropy of $S'$ minus the conditional entropy of $S'$ given $S$. 
It is large when $S'$ takes many values with uniform probability and $S'$ can be predicted well from knowing the value of $S$. 
One can imagine that diverse but coherent movement will have a high mutual information of time consecutive sensor readings. 
Hence this quantity is a very natural form of intrinsic reward for embodied systems. 
The mutual information $\MI(S;S')$ is also the Kullback-Leibler divergence from the joint distribution of $S$ and $S'$ to its best approximation by a factorizing distribution, and is sometimes regarded as a type of complexity measure.

\subsection{Mutual information intrinsic reward signal}
We consider an intrinsic reward signal of the form 
\begin{equation}
\rMIt =  
\sum_{s'} p_{t}(s'|s_{t-1}) \log \frac{p_{t}(s'|s_{t-1})}{p_{t}(s')},  
\label{eq:MIrew}
\end{equation}
where $p_{t}$ is the empirical distribution of sensor readings computed from the last $T'$ time steps, 
\begin{equation}
p_{t}(s,s') = \frac{1}{T'} \sum_{u=1}^{T'} \delta_{(s,s')}(s_{t-u},s_{t+1-u}), 
\quad 
p_{t}(s) = \frac{1}{T'} \sum_{u=1}^{T'} \delta_{s}(s_{t+1-u}). 
\end{equation}
With this definition, the time average $\RMI = \frac{1}{T}\sum_{t=1}^{T} \rMIt$ is an estimate of the mutual information $\MI(S;S')$. 
Here we can also compute $r_{\MI_i,t}$ for individual sensors $i=1,\ldots, N$, 
and use the average $r_{\MI,t} = \frac{1}{N}\sum_{i=1}^Nr_{\MI_i,t}$ as the intrinsic reward. 
This is high when all degrees of freedom move to positions that contribute to a high empirical entropy and at the same time are predictable from their previous values. 
The advantage is that the mutual information of individual sensors can be estimated accurately with a smaller $T'$. Furthermore, in contrast to the joint mutual information, maximizing this does not require that all joint sensor states are occupied with positive probability. 

The mutual information will inevitably vary during the learning process, 
and we are not compelled to estimating it more accurately than necessary nor optimizing for the highest possible value. 
It is worthwhile mentioning that, in stationary settings, uniform estimation error bounds~\citep{Shamir20102696} show that estimating the mutual information is much easier than estimating the joint distribution, with a complexity bound controlled by the number of sensor states that occur with positive probability.

\subsection{Combined reward signal}

Our ultimate goal is to learn a policy that produces a high value of some extrinsic reward, 
which in our running example is the average locomotion velocity of an embodied agent, $\RLoc = \frac{1}{T}\sum_t \rLoct$, where $\rLoct$ is the distance covered in the time between $t-1$ and $t$. 
Instead of optimizing $\RLoc$ by itself, 
we optimize a combined reward that includes the predictive information as an intrinsic reward. 
As the total reward signal $r_t$ for the agent we use a geometric mixture, 
which for a mixture weight $\xi\in[0,1]$ and non-negative components $\rLoct$ and $\rMIt$ is given by 
\begin{equation}
r_t = \rLoct^\xi \cdot \rMIt^{1-\xi} . 
\label{eq:combinedreward}
\end{equation}
For a better comparability, in practice we scale $\rMIt$ and $\rLoct$ so that both have approximately the same maximum possible value, and $r_t$ so that it has a maximum possible value of approximately~$1$. 
The signal can also be modulated with a monotonic function $f$ as $r_t = f(\rLoct^\xi \cdot \rMIt^{1-\xi})$. For instance, taking $f$ as the square function dampens low values and amplifies high values. 
In order to account for negative locomotion rewards, we can simply set $r_t = \operatorname{sign}(\rLoct)\cdot f(|\rLoct|^\xi \cdot \rMIt^{1-\xi} )$. 
Combining objective functions by building their product is a standard method to avoid that during learning only one of them is optimized. 
The parameter $\xi$ allows us to place more or less importance on one or the other component. 
Other types of combinations are possible, 
and in general the type of combination will have an effect on the optimization problem.

\section{Related work}
\label{sec:relatedwork}

There is a large body of work related to intrinsic motivation and information theory in the context of learning to act. Here we can only comment on a few papers and refer the reader to the references provided therein. 
In its core, intrinsic motivation seeks to define a reward signal when there is no specific task objective~\citep{Barto2013}. 
\cite{NIPS2004_2552} proposed intrinsically motivated reinforcement learning as a framework to allow agents learn hierarchical collections of skills autonomously, and tested it in artificial playrooms. 
\cite{Steels2004} studied the balance of skill and challenge of behavioral components as the motivation for open ended development of embodied agents. 

Intrinsic motivation has drawn much attention as a tool for improving exploration and the ability to make predictions. 
\cite{Rubin2012} studied curiosity driven intrinsic motivation as an exploration incentive. 
\cite{4141061} used a form of prediction error as a reinforcement signal. 
\cite{10.3389/fncir.2013.00037} studied the predicted information gain in model based reinforcement learning as a way to encourage actions that yield most information about the structure of the world. 
\cite{Schmidhuber2009} considered a compression quantity as a reinforcement signal, arguing that data is temporarily interesting by itself once the agent learns to predict or compress it. 
\cite{10.3389/fnbot.2013.00025} proposed curiosity as an intrinsic reward to encourage actions taking the agent to regions where it can learn something about the world, and studied this approach in complex systems with reactive policies. 

Two information theoretic quantities that have been studied intensively in the context of learning are empowerment and predictive information. 
Empowerment~\citep{1554676} is the maximal information of an action sequence about a future state, 
and assigns a value to each possible initial state. 
Maximizing it will encourage the agent to occupy positions from which it can reach most states within its planning horizon. 
\cite{NIPS2015_5668} studied the efficient computation and maximization of empowerment as a learning principle in maze environments. 
The empowerment has been investigated as an incentive to have an agent actively structure his environment in relation to his embodiment~\citep{e16052789}. 
\cite{Still2012} stressed the importance of choosing policies that not only maximize a task objective, 
but also allow for high predictive power, which should make the world both interesting and exploitable. 
\cite{Zahedi:2010:HCL:1841970.1841973} studied the predictive information maximization as a learning principle, and demonstrated that it can generate coordinated behavior in embodied agents. 

Probably closest to our investigations are the following three works. 
\cite{Prokopenko2006aEvolving} used the predictive information, estimated on the spatio-temporal phase-space of an embodied system, as part of the fitness function in an artificial evolution setting. 
It was shown that the resulting locomotion behavior of a snake-bot was more robust, compared to the setting, in which only the traveled distance determined the fitness. 
\cite{10.3389/fpsyg.2013.00801} studied the predictive information as a supplement to an extrinsic task related reward. Like us they studied embodied systems, but they considered episodic tasks, used linear combinations of the intrinsic and extrinsic rewards, deterministic controllers, and stochastic policy search. Although initially beneficial, asymptotically the predictive information led to weaker results. 
Consequently they suggested that different ways of combining the rewards and gradient based optimization should be investigated, which is the approach that we take here. 
\cite{e18010006} studied a broad range of information theoretic quantities, including the predictive information, for aiding the optimization of task objectives. In the same spirit as us, they suggested that intrinsic rewards could help surmount optimization barriers. 
Like us, they used multiplicative combinations, but for the predictive information they obtained sobering results, showing no substantial benefit over direct optimization and other information theoretic quantities. In contrast to us, however, they considered deterministic policies optimized by evolutionary algorithms. Furthermore, they focused on a particular task within a 2-D world that is relatively simple in comparison to our physically realistic embodied system. 
While these previous works used episodic or evolutionary methods, 
here we use a model-free online policy gradient optimization method.

\section{Experiments}
\label{sec:experiments}

\subsection{Svenja virtual robot}

\begin{figure}
\centering
\scalebox{.9}{
\begin{tabular}{c}
\includegraphics[clip=true, trim=7cm 0cm 8cm 0cm, height=5.1cm]{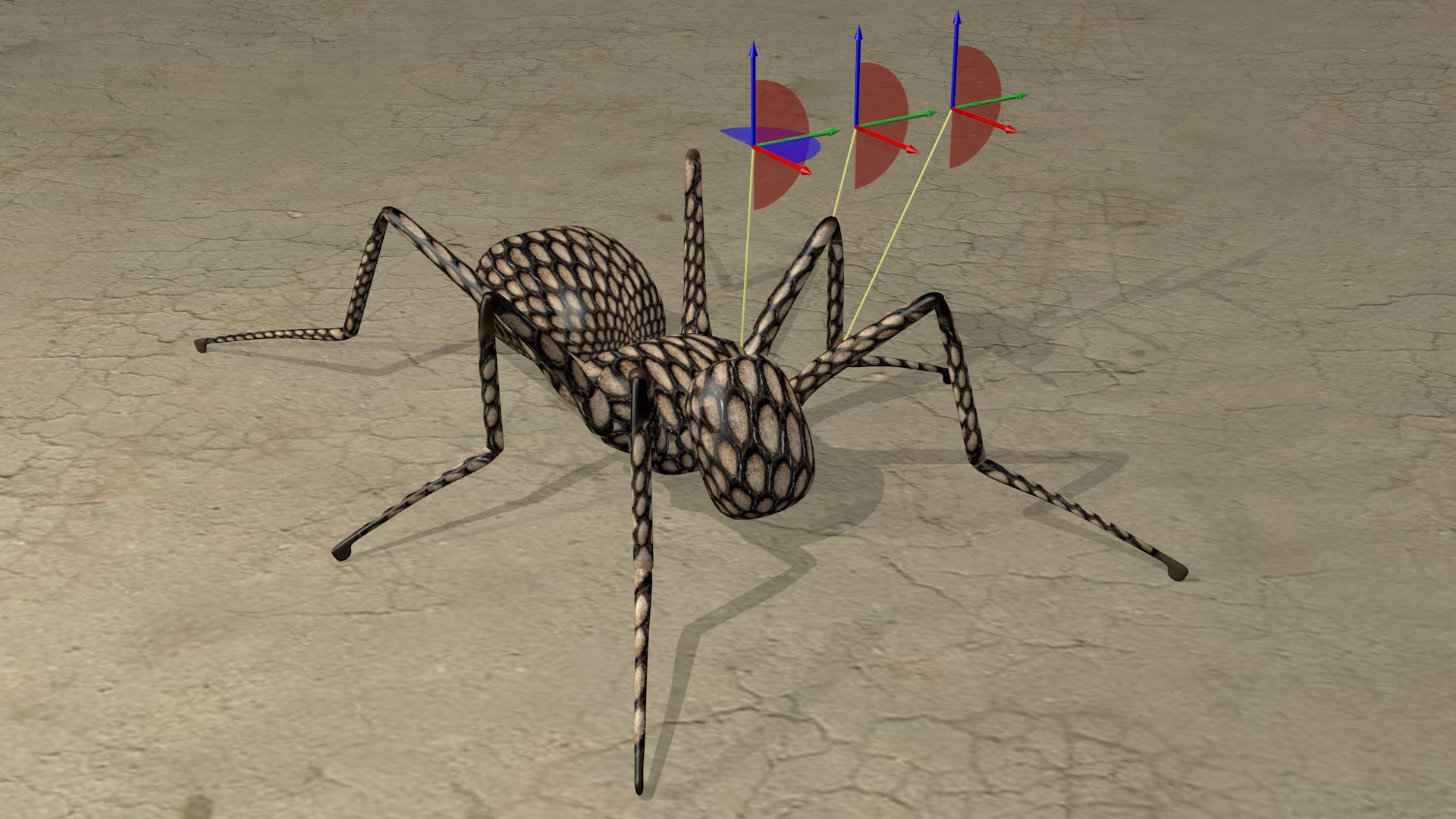}
\end{tabular}
\hspace{-.5cm}
\begin{tabular}{c}
 \includegraphics[clip=true, trim=5cm 0cm 7cm 0cm, height=2.5cm]{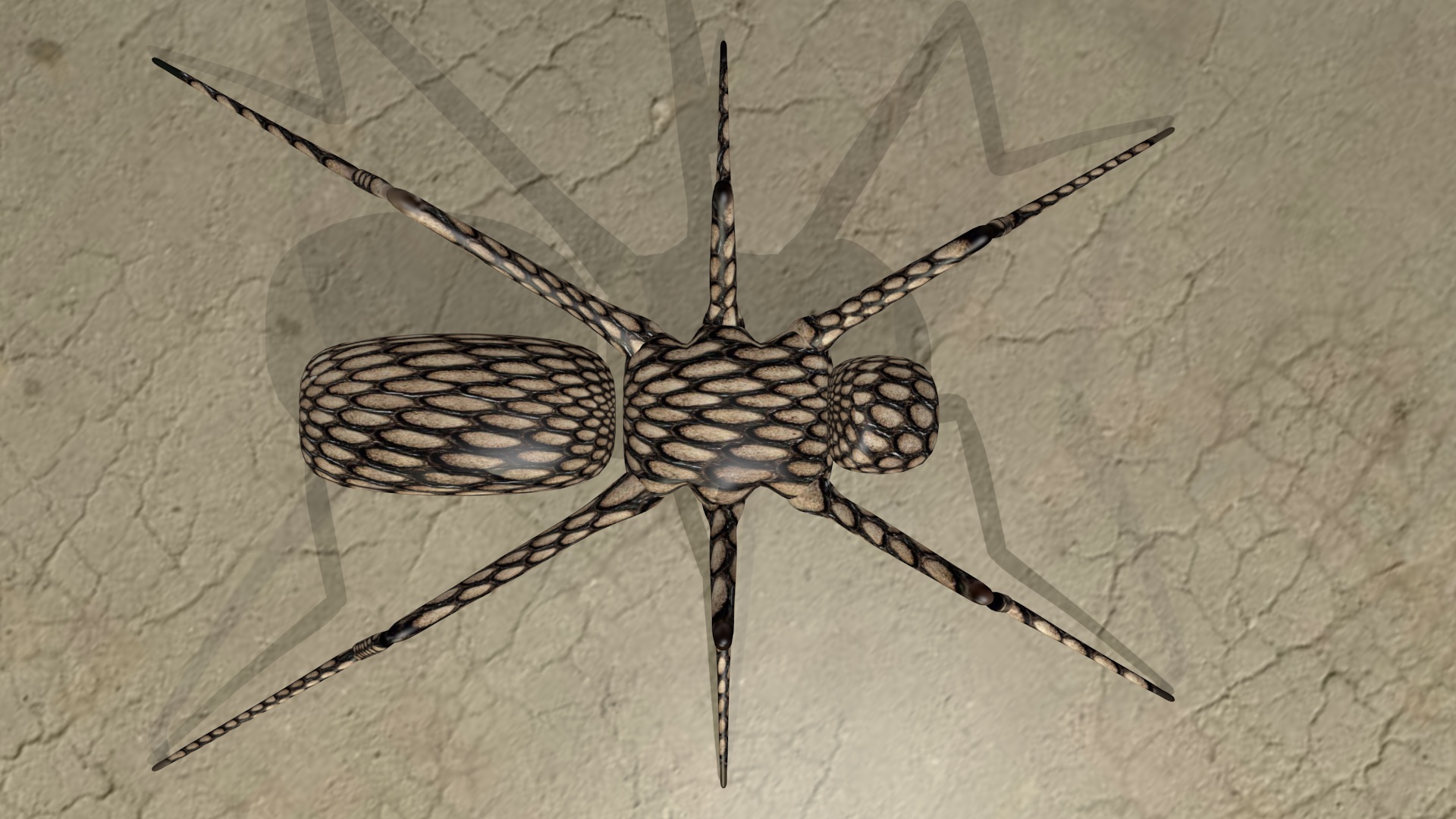}\\
\includegraphics[clip=true, trim=5cm 0cm 7cm 0cm, height=2.5cm]{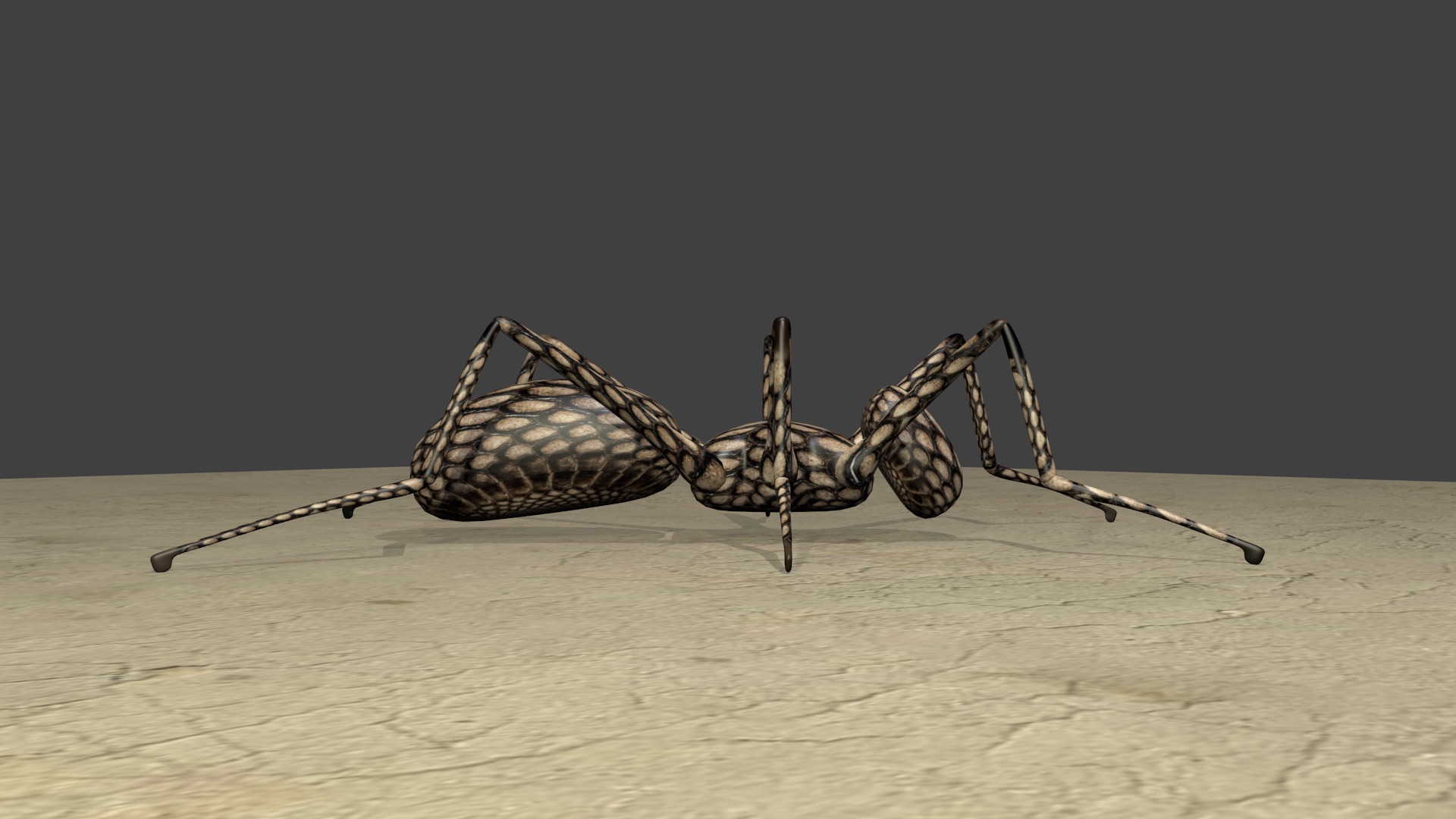}
\end{tabular}
}
\caption{Svenja virtual hexapod. 
Each leg has four controllable continuous degrees of freedom with corresponding sensors and a binary foot contact sensor. 
Joints at the main body attachment are endowed with damped springs. 
	}
	\label{fig:sfiantpicture}
\end{figure}

Svenja is a complex embodied agent, shown in Figure~\ref{fig:sfiantpicture}, inspired by the morphology of an ant. 
It is a hexapod with $3$ actuated joints in each leg: a spring spherical joint with two degrees of freedom and two knees with one revolute joint each. 
This makes a total of $24$ controllable degrees of freedom. 
It has $24$ continuous sensors, measuring the joint positions, and $6$ binary foot contact sensors. 
The body is bilaterally symmetric, but the three legs on each side are different from each other in length, weight, orientation, and maximum torques of the joints. 
Svenja is simulated in YARS~\citep{zahedi2008yars}, which uses the physics engine bullet. 

\subsection{The learning task}
As the extrinsic locomotion reward we consider the average forward velocity, 
$\RLoc = \frac{1}{T}\sum_{t=1}^T \rLoct$, 
where $\rLoct$ is the distance covered in the forward direction in the last time step, 
\begin{equation}
\rLoct = 
(\pos_{t} - \pos_{t-1} )^\top d_{t}, 
\end{equation}
$\pos_t$ is the position of the center of gravity in the XY-plane and $d_t$ is the XY-part of the axial direction of the robot. 
Training was set up to optimize the time average of the combined signal $r_t = \operatorname{sign}(\rLoct)(|\rLoct|^\xi\cdot\rMIt^{(1-\xi)})^2$, with an intrinsic reward $\rMIt = \frac{1}{N}\sum_ir_{\MI_i,t}$ corresponding to the mean mutual information per sensor.

\subsection{The policy model}

As a controller architecture we use the conditional restricted Boltzmann machine (CRBM), 
which is a parametric model of Markov kernels of the form 
\begin{equation}
p_\theta(y|x) = \frac{1}{Z_\theta(x)} \sum_{h} \exp(h^\top V x + h^\top W y + c^\top h + b^\top y), 
\end{equation}
where $x\in\{0,1\}^k$ are input binary vectors, $y\in\{0,1\}^n$ are output binary vectors, $h\in\{0,1\}^m$ are hidden binary vectors, $\theta=\{V,c,W,b\}$ are real valued matrices of parameters, and $Z_\theta$ is a normalizing partition function. 
We consider reactive policies, where $x$ corresponds to the current sensor state $s_t$ and $y$ to the action $a_t$. 
If desired, one can also implement policies with memory by taking temporal sensor sequences $(s_{t-k},\ldots, s_t)$ for $x$. 

Computing the partition function $Z_\theta(x)$ is intractable. 
The typical sampling procedure is to run a short sequence of Gibbs updates, with fixed $x$, and then return $y$. 
Since there are only connections between visible and hidden units, 
all entries of $h$ and all entries of $y$ can be updated in parallel. 
The gradient $\nabla_\theta \log p_\theta(y|x)$ is also intractable. 
It is approximated using Monte Carlo averages, with the sampling procedure described above. 
This method is widely used for generative training of restricted Boltzmann machines and is familiar from the Contrastive Divergence algorithm~\citep{HintonAParactical}.

\subsection{The learning algorithm}

We estimate the gradient of the average reward $R$ with respect to the policy parameters $\theta$ in a simulation-based manner using the GPOMDP algorithm~\citep{Baxter:2001:IPE:1622845.1622855}. 
At simulation times $t$ with $\operatorname{mod}(t,T)=0$, for some fixed $T\in\mathbb{N}$,  
we start a gradient estimation loop by initializing auxiliary variables $z_{t}=0$, $\Delta_{t}=0$, $\bar r_{t} = 0$. These variables are then updated by 
\begin{align}
z_{t+1} =& \beta_{\text{GPOMDP}} \cdot z_t + \nabla_\theta \log p_\theta(a_t|s_t) \\
\Delta_{t+1} =& \Delta_t + (r_{t+1}\cdot z_{t+1} - \Delta_t ) / \operatorname{mod}(t+1,T) \\
\bar r_{t+1} =& \bar r_t + r_t . 
\end{align}
Here $\beta_{\text{GPOMDP}}$ is a discount parameter between zero and one, 
which, in combination with $T$, allows to control a bias variance tradeoff. 
The estimation loop ends after $T$ controller iterations, when we read out the estimate $\Delta = \Delta_t$ of the average reward gradient and the average reward $R = \bar r_t/T$. 
With the gradient estimate at hand, we update the policy parameter $\theta$ by  
\begin{align}
\Delta_\theta \leftarrow & \alpha_{\text{learn}}\cdot ( \Delta  - \alpha_{\text{decay}}\cdot \theta) + \alpha_{\text{mom}}\cdot \Delta_{\theta} \\ 
\theta \leftarrow&  \theta + \Delta_\theta. 
\end{align}
Here $\alpha_{\text{learn}}$ is a positive learning rate, 
$\alpha_{\text{decay}}$ is a weight decay parameter between zero and one, 
and $\alpha_{\text{mom}}$ is a non-negative momentum parameter. 
For the very first update $\Delta_\theta$ is initialized with zeros. 

\subsection{Experimental setup}

For this experiment we fixed the physics simulator frequency to $100$~Hz, with $200$ solver iterations per physics update, and fixed the controller frequency to $10$~Hz. 
As inputs to the CRBM controller we used binarized versions of all sensor measurements at the current time step. 
The $24$ continuous sensor values were each reduced to one bit and the $6$ foot contact sensors were left untouched, making a total of $30$ input units. 
The CRBM binary output vectors were mapped to the bin centroids as actuator states and passed back to the simulator. 
Here again we used one bit per actuator, making a total of $24$ output units. 
We used a fully connected CRBM with $50$ hidden units. 

For each training instance, each parameter of the CRBM was initialized as $0.01$ of a normally distributed random sample. 
We fixed learning rate $\alpha_{\text{learn}}=1$, momentum $\alpha_{\text{mom}}=0.5$, and weight decay $\alpha_{\text{decay}}=10^{-5}$. 
We used $10$ full Gibbs updates initialized at random for generating each output vector and $60$ samples for each Monte Carlo average. 
We fixed the GPOMDP discount parameter $\beta_{\text{GPOMDP}}=0.75$ and used $T=100$ controller iterations for each gradient and average reward estimate. 
The intrinsic reward was computed using the previous $T'=100$ sensor measurements. 

Every $1000$ controller iterations, the next $10$ controller outputs were corrupted with a small amount of noise. 
This corresponds to an external force acting on the body of the agent from time to time, like a strong wind, and forces the agent to learn more robust behaviors. 
We also included a small negative reward (penalty) for touching the floor with the head or with the rear. 

In order to ensure that our results were representative, we run $30$ training instances for each choice $\xi=0,0.25,0.5,0.75,1$ of the reward mixture weight. 
Each instance was interrupted after the same fixed computation time, corresponding to about $30,\!000$ gradient updates and $3,\!000,\!000$ controller iterations. 

\begin{figure}
	\centering
	\scalebox{.9}{
		\includegraphics[clip=true,trim=4.5cm 19cm 6cm 18cm,width=\textwidth]{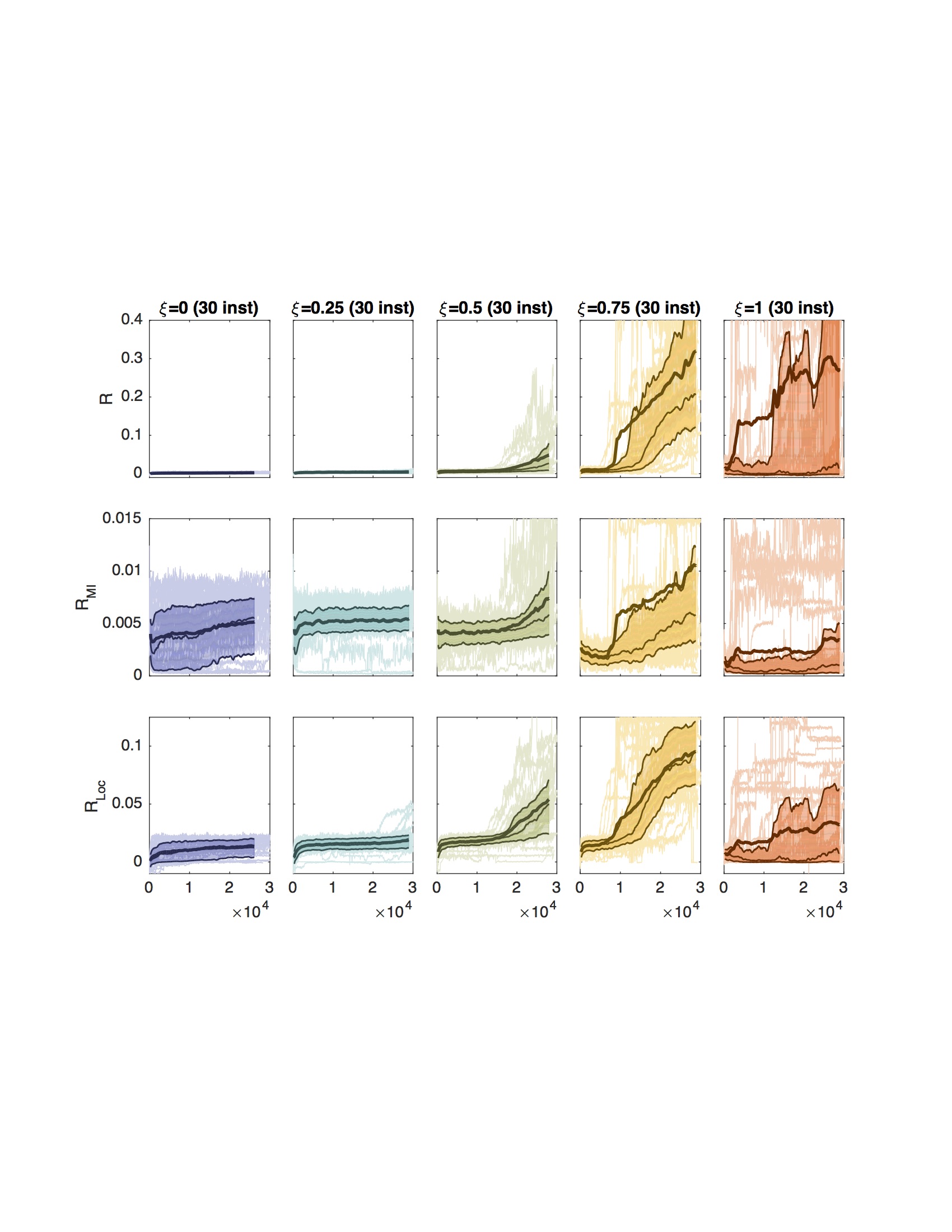}
	}
	\caption{Experimental results on Svenja. 
		Shown are the values of the combined reward $R$, the intrinsic reward $\RMI$, and the extrinsic reward $\RLoc$, over the policy parameter updates (with $1$ parameter update every $10$ Svenja time seconds). 
		The columns correspond to mixture weights $\xi = 0,0.25,0.5,0.75,1$. 
		The dark thin lines show the first quartile, median, and third quartile for the $30$ training instances. 
		The area enclosed by these lines contains $50\%$ of the data. 
		The thick lines show the mean across training instances. 
	}
	\label{fig:sfiant}
\end{figure}

\subsection{Results}
\label{sec:results}

Figure~\ref{fig:sfiant} shows the evolution of $R$, $\RMI$, and $\RLoc$ over the number of gradient updates, for the different mixture weights tested. 
Training with $\xi=0.75$ produced the highest values of $\RLoc$, 
in terms of the maximum, mean, and median across training instances. 
Training with $\xi=0.5$ also produced good results, but strong improvements set on later and were still progressing at the moment that training was interrupted. 
Training with $\xi=0$ and $\xi=0.25$ did not yield high locomotion values, which is not surprising, since the reward signal emphasized mainly the mutual information. 
Training with $\xi=1$, which corresponds to optimizing $\RLoc$ by itself, often led to high values of $\RLoc$, 
but more than half the instances did not make any significant progress, 
and on average the performance was relatively poor.

\begin{figure}
	\centering
	\scalebox{.9}{
	\includegraphics[clip=true,trim=15cm 29cm 17cm 29cm, width=.6\textwidth]{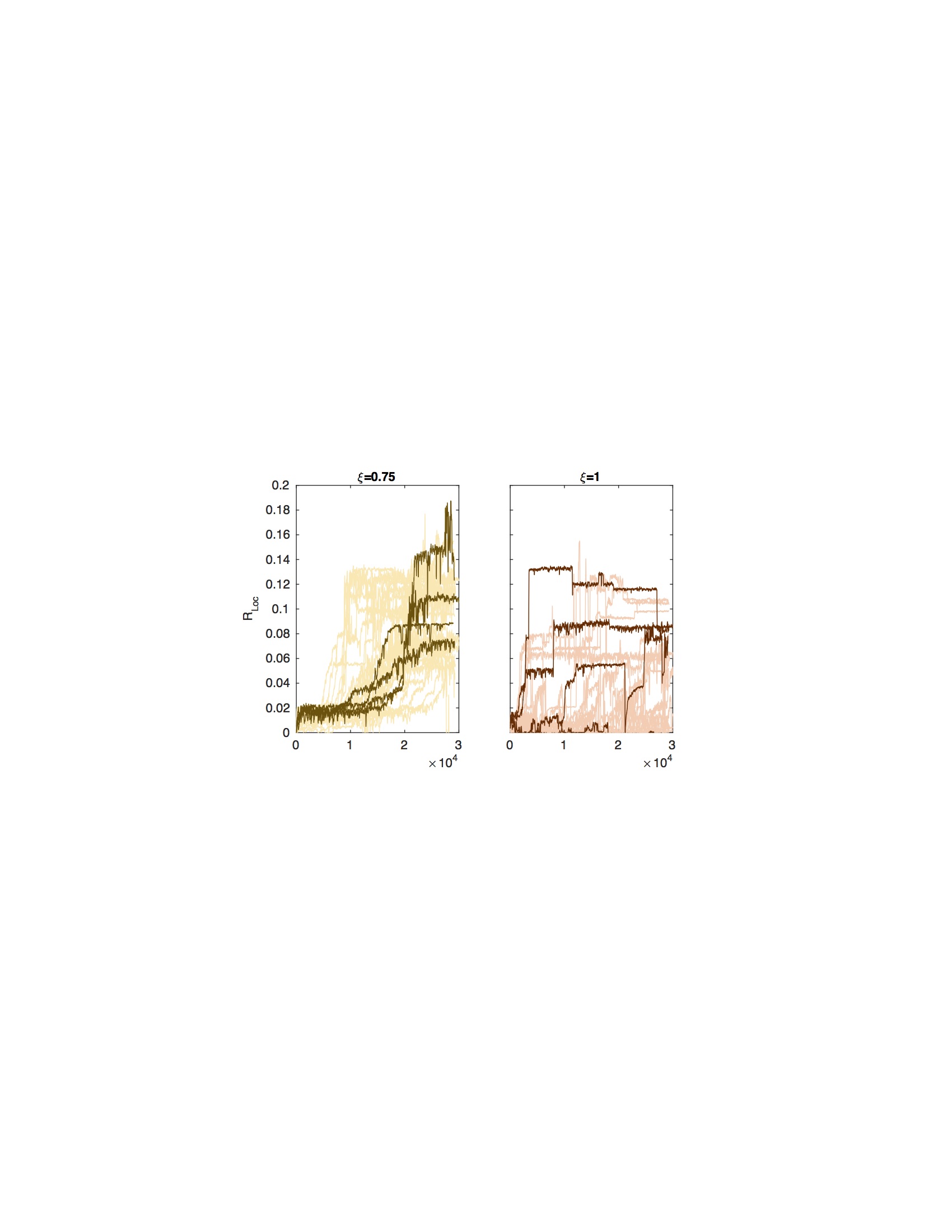} 
	\quad 
	\includegraphics[]{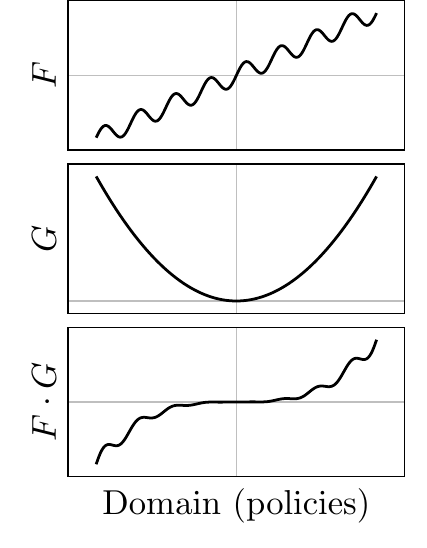}
	}
	\caption{
		Optimization landscapes with and without mutual information. 
		The two plots to the left are a close up of Figure~\ref{fig:sfiant}, 
		showing $R_{\text{Loc}}$ over the number of gradient iterations for $\xi=0.75$ and $\xi=1$. 
		A few representative training instances are highlighted. 
		The right panel sketches of our interpretation: for an appropriate pair of functions $F$ and $G$, local optimizers of $F$ are smoothed out after multiplication with $G$, while concurrent global optimizers of both functions are preserved.  
	}
	\label{fig:sfiant_detail}
\end{figure}

\begin{figure}
	\centering
	\scalebox{.9}{
	\includegraphics[clip=true,trim=2.5cm 10.5cm 2.5cm 10cm, width=\textwidth]{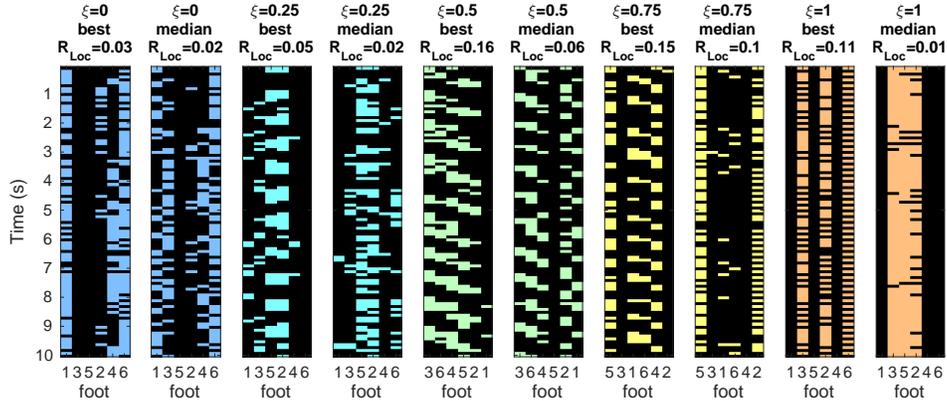}
	}
	\caption{Here we show sequences of foot contact readings for the instances with the highest and median value of $\RLoc$ a few moments before training was interrupted, for $\xi=0, 0.25, 0.5, 0.75, 1$. Light color indicates foot contact. The foot enumeration is 1 front left, 2 front right, 3 middle left, 4 middle right, 5 rear left, 6 rear right. 
	}
	\label{fig:footspart}
\end{figure}

Figure~\ref{fig:sfiant_detail} shows that $\RLoc$ improves much more smoothly when including a certain amount of mutual information in the optimization objective ($\xi=0.75$, meaning $25\%$ mutual information). 
With pure locomotion reward ($\xi=1$), some training instances show quick improvements and arrive at quite good solutions. 
Nevertheless, very often they get trapped in local optimizers, 
and sometimes the performance drops sharply, 
indicating that the learned policies are not robust. 
Our interpretation for this behavior is that the combination with the mutual information smoothens out local optimizers of the locomotion reward while preserving its global optimizers, 
as illustrated in the right panel of the figure. 
This explanation is well in agreement with the intuition, also observed here, that behaviors with high values of $\RLoc$ tend to first traverse or be accompanied by high values of $\RMI$. 
This is also in agreement with the idea that maximizing the mutual information allows for a better exploration of sensor values, which naturally increases the robustness of the policies. 
Figure~\ref{fig:footspart} shows a few sequences of foot contact sensor for the different mixture weights at the moment that training was interrupted. 
We include videos in the Supplementary Material.

\subsection{Other configurations}

We tested various alternative training configurations, taking more bits per continuous sensor and actuator, different numbers of hidden units, and various types of restricted connectivity structures (e.g., one individual block of CRBM hidden units per actuator). 
We also run experiments using the mean mutual information over the joint sensor states of each leg, a version of Svenja with constrained joint mobility, and different types of combinations of intrinsic and extrinsic rewards. 
In all cases the results were similar to the ones that we report here, showing that a moderate amount of mutual information intrinsic reward significantly improves the learning performance. 
We include some of these results in the Supplementary Material. 

\section{Conclusion}
\label{sec:conclusions}

Our experiments show that for complex embodied systems, 
combining the extrinsic reward for a locomotion task with a judicious amount of intrinsic reward, 
defined as the predictive information of time consecutive sensor readings, 
can significantly improve the policy gradient optimization process. 
Our observation is that combining extrinsic and intrinsic reward signals can considerably smoothen the optimization landscape while preserving global optimizers of the extrinsic reward. 

We think that for a broad range of learning tasks, especially involving coordinated movement of embodied systems, the predictive information is a useful form of intrinsic reward that can aid learning. 
This paper demonstrates that substantial benefits can be gained from this approach. 
A deeper analysis of intrinsic rewards as smootheners of the optimization landscape is certainly a promising avenue of research, searching for optimal ways of combining functions and the best types of landscape regularizers for particular collections of tasks. 

Of course, we are also interested and are currently working on experiments with other types of morphologies and tasks. 
Also, we are interested in the behaviors that can emerge from optimizing the mutual information of longer temporal sequences of joint sensor values (considering more than two consecutive time steps), and scheduled combinations of intrinsic and extrinsic rewards, where the intrinsic reward is turned down after an initial stage in order to fine tune for the extrinsic reward. 
In the past, investigations of this type have been limited to much simpler morphologies. We think that the main reason was that the optimization approaches either involved random search, did not have scalable policy models, or required explicit inference of the world state transition probabilities, all difficulties that are sorted out in our approach. 

Another interesting practical aspect is model selection. We did not divert much in this direction here, but it is clear that an appropriate choice of the policy model can have a significant effect on the optimization problem. In particular, the morphology can be better accounted for by using suitable modular controller architectures.

\FloatBarrier

\renewcommand*{\bibfont}{\footnotesize}
\bibliography{referenzen}
\bibliographystyle{abbrvnat}

\newpage

\appendix

\section*{Supplementary Material}

Here we show the learning curves for various experiments that we conducted on the robot Svenja. 
We varied the joint mobility of Svenja, the form of the mutual information reward, the number of bits in the discretization of continuous degrees of freedom, and the scaling of the combined reward. 
All experiments produced similar outcomes to those presented in the main part. 
Including a certain amount of mutual information intrinsic reward consistently led to significant improvements in the locomotion performance.

\begin{figure}[h]
	\centering
	\includegraphics[clip=true,trim=4.5cm 19cm 6cm 18cm,width=\textwidth]{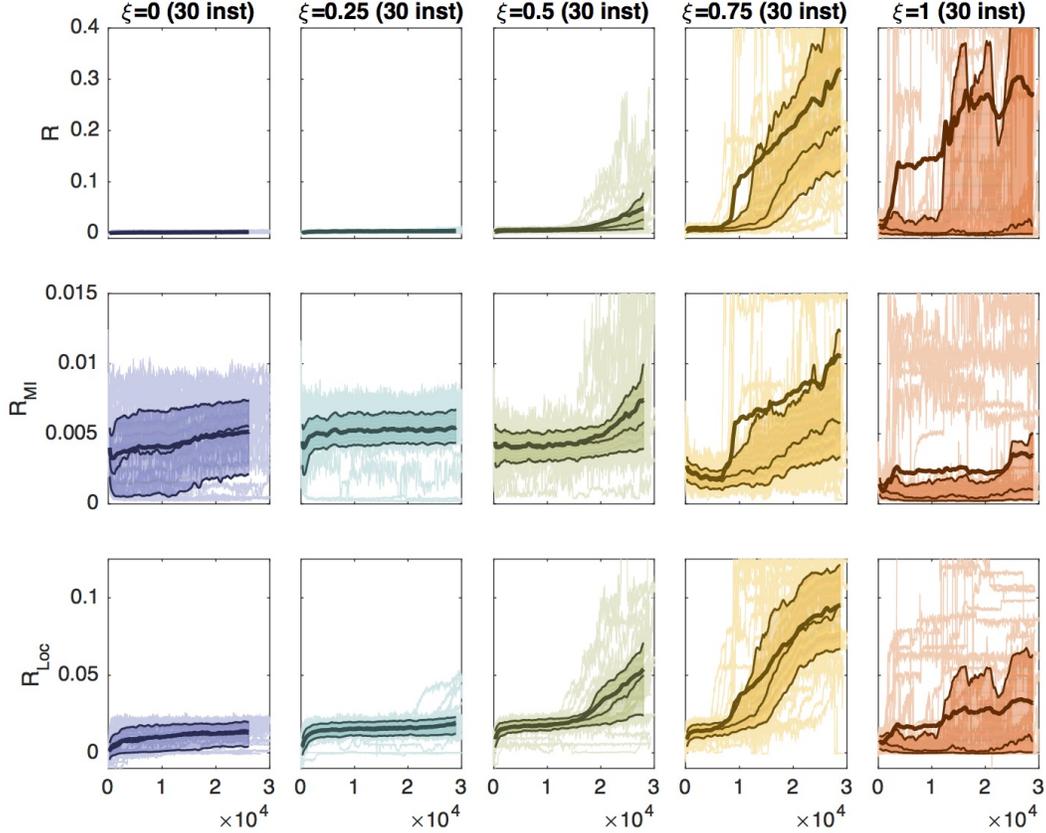}
	\caption{Experimental results on Svenja. 
		Here we set $\rMIt$ equal to the average of the mutual information per degree of freedom, scaled to have a maximum possible value of $0.1$, 
		one bit per degree of freedom, and optimized the time average of $r_{t} = \operatorname{sign}(\rLoct)\cdot (|\rLoct|^\xi \cdot \rMIt^{1-\xi})^2$. 
		Shown are the policy gradient learning curves over the number of gradient iterations. 
		The columns correspond to $\xi = 0,0.25,0.5,0.75,1$ from left to right, 
		for the indicated number of instances of the experiment. 
		The first row shows the reward per time step. 
		The second row shows the average mutual information per sensor. 
		The third row shows the average forward velocity. 
		The dark thin lines show the first quartile, median, and third quartile for all experiment instances. 
		The area enclosed by these lines contains $50\%$ of the data. 
		The thick lines show the mean values across instances of the experiments. 
		This figure is also contained in the main part of the paper and included here for completeness. 
	}
	\label{fig:sfiant}
\end{figure}

\begin{figure}
	\centering
	\includegraphics[clip=true,trim=4.5cm 31cm 6cm 32cm,width=\textwidth]{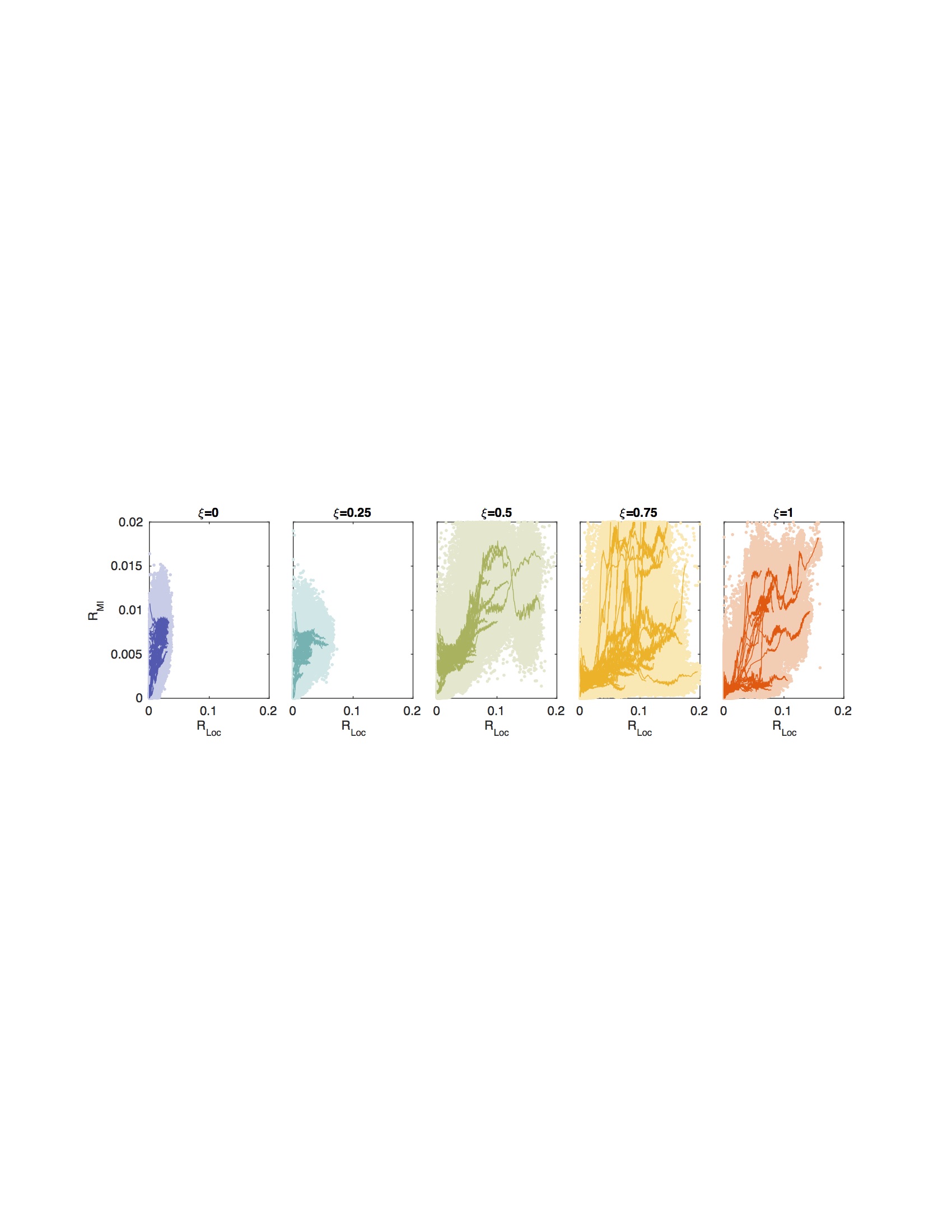}
	\caption{
		This figure plots the mutual information as a function of the locomotion reward for the data from Figure~\ref{fig:sfiant}. The curves show a sliding window average for individual training instances.  
		Higher locomotion values are often accompanied by higher mutual information values. 
		The sample Pearson correlation coefficient computed over all data is $0.7888$. 
	}
	\label{fig:sfiantMIvsLoc}
\end{figure}

\begin{figure}
	\centering
	\includegraphics[clip=true,trim=2.5cm 5cm 2.5cm 4cm, width=\textwidth]{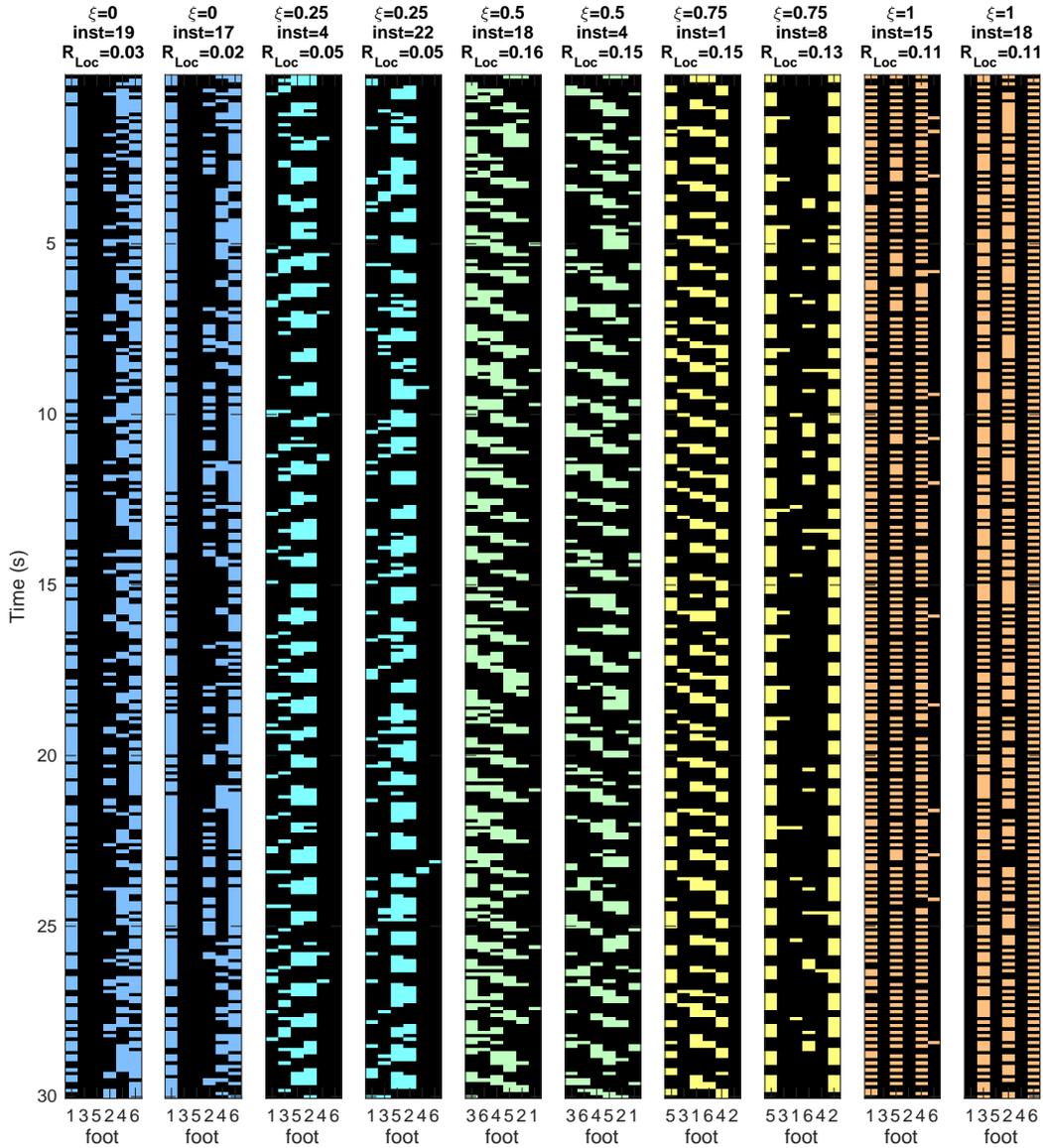}
	\caption{
		This figure illustrates the gaits that Svenja learned in the experiments from Figure~\ref{fig:sfiant}. 
		Shown are foot contact sequences for the two instances with the highest value of $\RLoc$, for $\xi=0, 0.25, 0.5, 0.75, 1$, 
		a few moments before training was interrupted. 
		Light color indicates foot contact. 
		The feet are enumerated 1 front left, 2 front right, 3 middle left, 4 middle right, 5 rear left, 6 rear right. }
	\label{fig:foots}
\end{figure}

\begin{figure}
	\centering	
	\includegraphics[clip=true,trim=4.5cm 19cm 6cm 18cm,width=\textwidth]{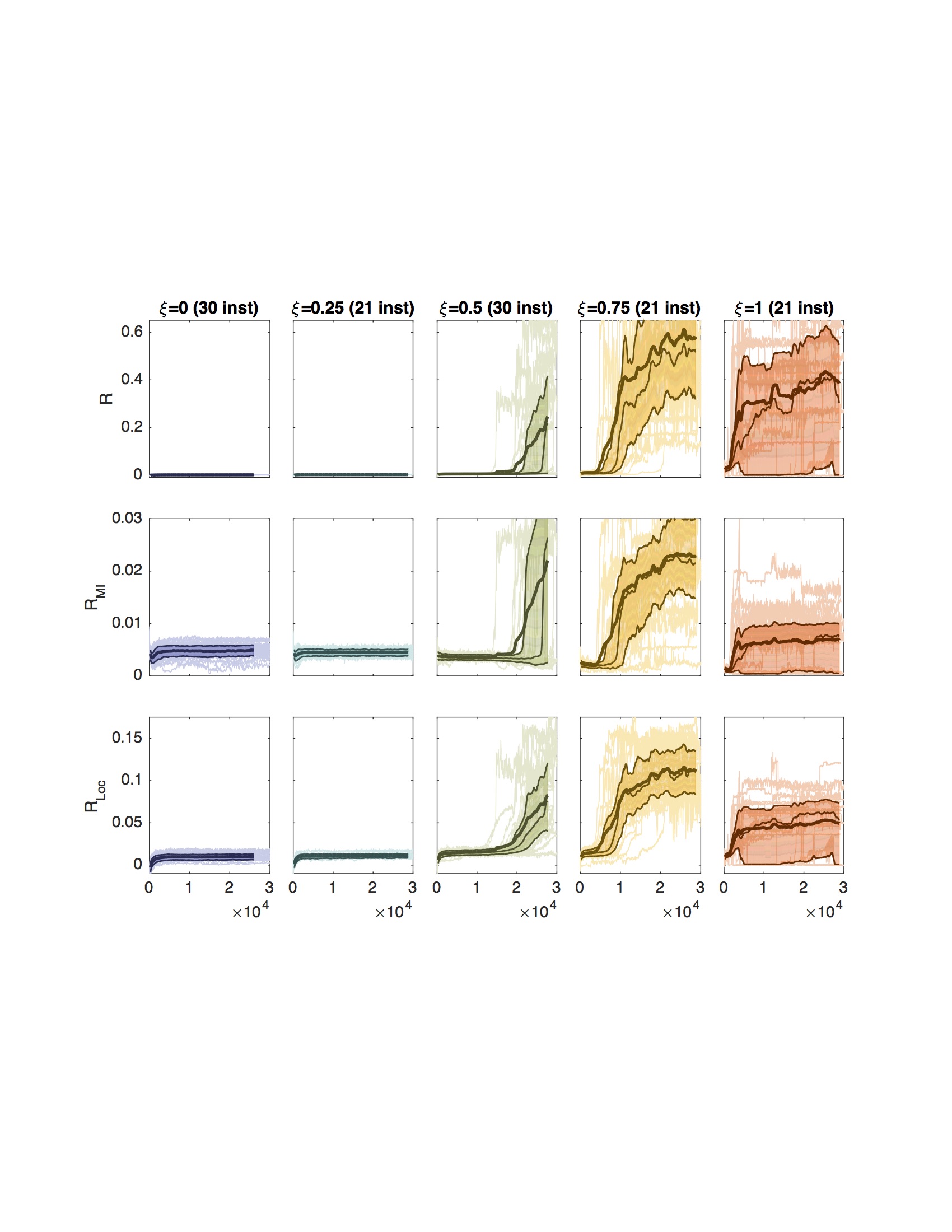}
	\caption{Experimental results on a modified version of Svenja. 
		This modification had constrained joint movement and was lighter and weaker than the original Svenja. 
		Here we set $\rMIt$ equal to the average of the mutual information per degree of freedom, 
		scaled to have a maximum possible value of $0.1$, 
		one bit per degree of freedom, 
		and optimized the time average of $r_{t} = \operatorname{sign}(\rLoct)\cdot (|\rLoct|^\xi \cdot \rMIt^{1-\xi})^2$. 
		The other details are as in Figure~\ref{fig:sfiant}. 
	}
	\label{fig:sfiant2}
\end{figure}

\begin{figure}
	\centering
	\includegraphics[clip=true,trim=2.5cm 5cm 2.5cm 4cm, width=\textwidth]{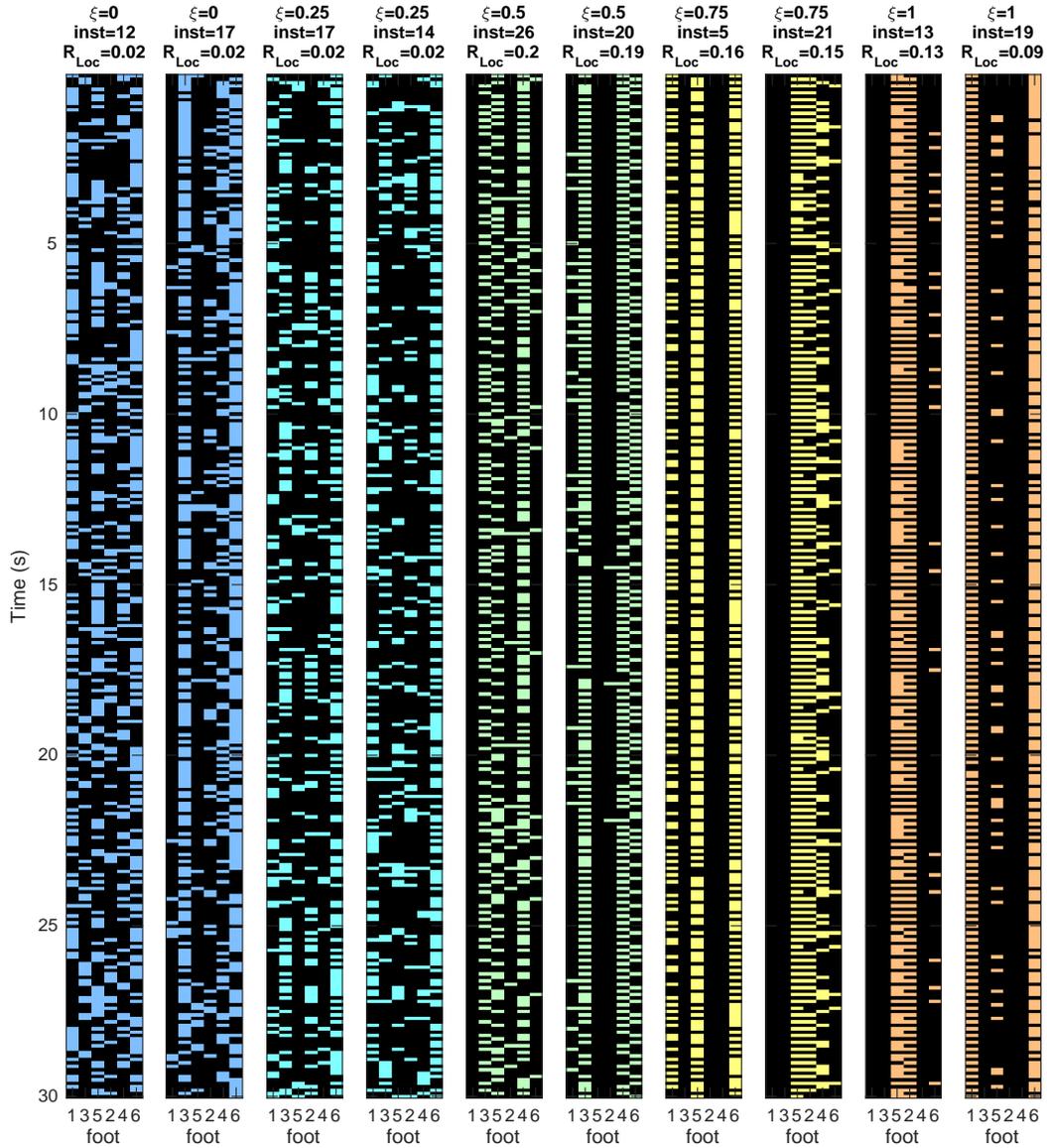}
	\caption{
		This figure illustrates the gaits that the modified version of Svenja learned in the experiments from Figure~\ref{fig:sfiant2}. 
		The modified version of the robot had constrained joint mobility and was lighter and weaker than the original. 
		The other details are as in Figure~\ref{fig:foots}. 
	}
	\label{fig:foots2}
\end{figure}

\begin{figure}
	\centering	
	\includegraphics[clip=true,trim=4.5cm 19cm 6cm 18cm,width=\textwidth]{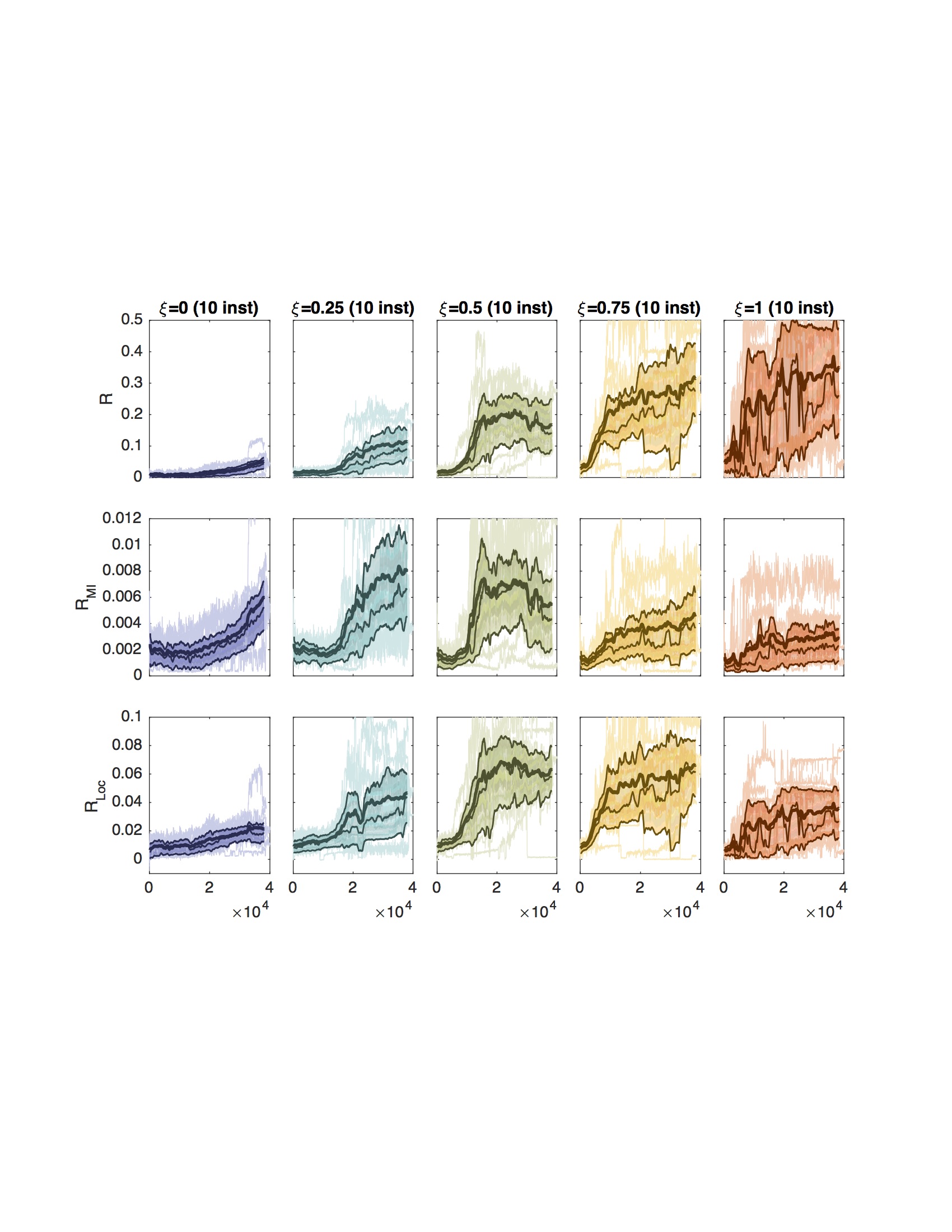}
	\caption{Experimental results on Svenja. 
		Here we set $\rMIt$ equal to the average of the mutual information per entire leg, 
		scaled to have a maximum possible value of $0.1$, 
		one bit per degree of freedom, 
		and optimized the time average of $r_{t} = \operatorname{sign}(\rLoct)\cdot |\rLoct|^\xi \cdot \rMIt^{1-\xi}$. 
		The other details are as in Figure~\ref{fig:sfiant}. 
	}
	\label{fig:sfiant2gr4nb1}
\end{figure}

\begin{figure}
	\centering	
	\includegraphics[clip=true,trim=4.5cm 19cm 6cm 18cm,width=\textwidth]{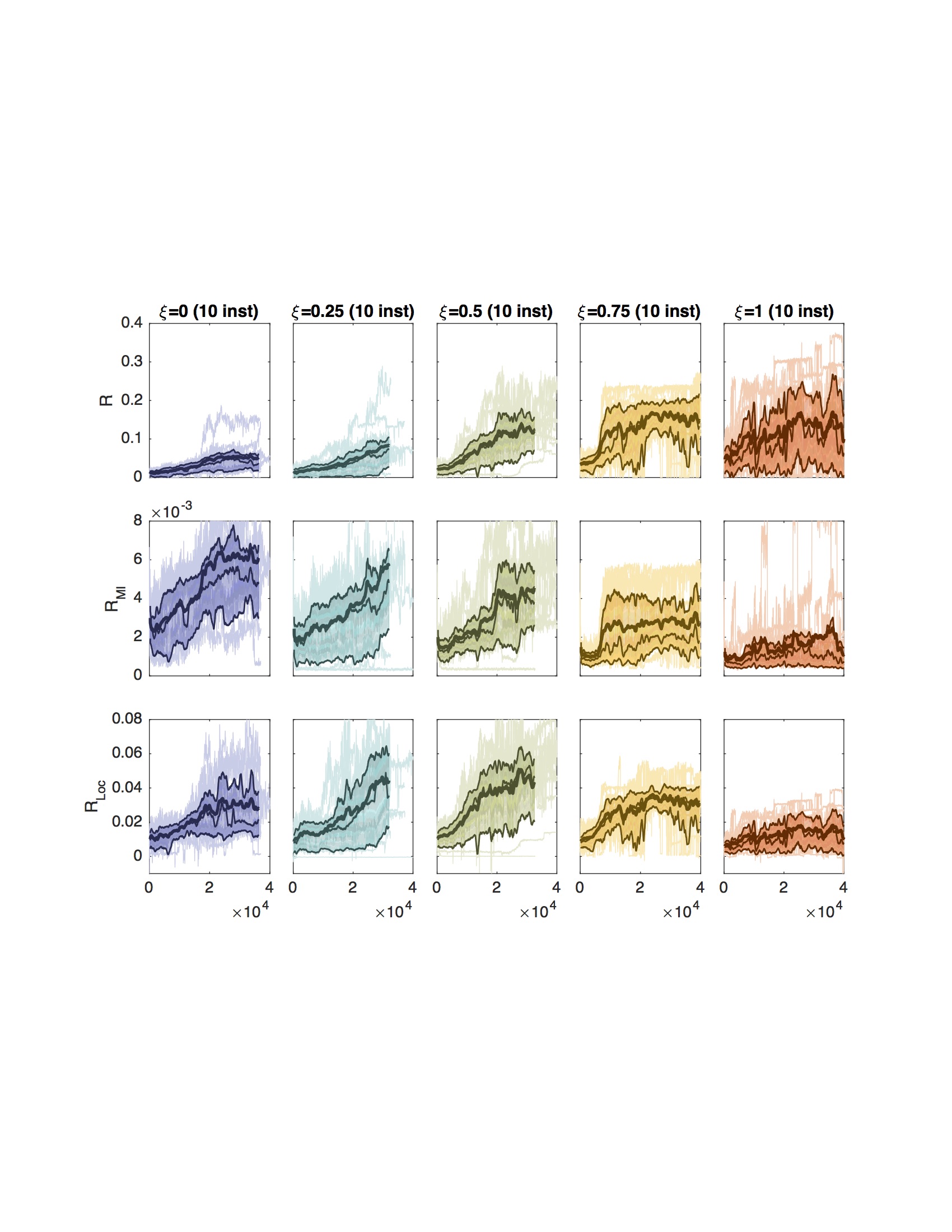}
	\caption{Experimental results on Svenja. 
		Here we set $\rMIt$ equal to the average of the mutual information per entire leg, scaled to have a maximum possible value of $0.1$, 
		two bits per continuous degree of freedom, 
		and optimized the time average of $r_{t} = \operatorname{sign}(\rLoct)\cdot |\rLoct|^\xi \cdot \rMIt^{1-\xi}$. 
		The other details are as in Figure~\ref{fig:sfiant}. 
	}
	\label{fig:sfiant2gr4nb2}
\end{figure}

\begin{figure}
	\centering	
	\includegraphics[clip=true,trim=4.5cm 19cm 6cm 18cm,width=\textwidth]{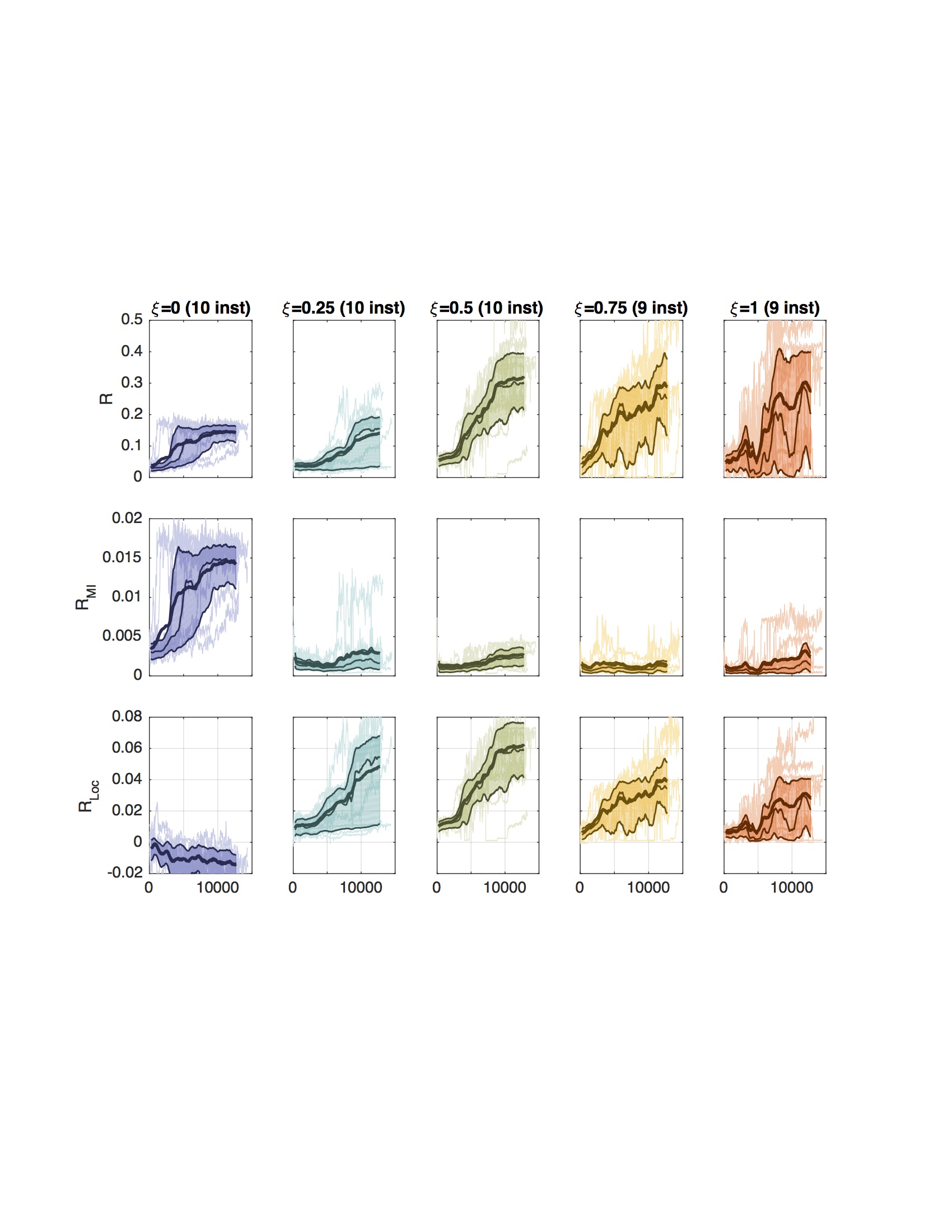}
	\caption{Experimental results on Svenja. 
		Here we set $\rMIt$ equal to the average of the mutual information per sensor, scaled to have a maximum possible value of $0.1$, 
		one bit per continuous degree of freedom, 
		and optimized the time average of $r_{t} = \xi \cdot \rLoct + (1-\xi)\cdot \rMIt$. 
		The other details are as in Figure~\ref{fig:sfiant}. 
	}
	\label{fig:linsfiant}
\end{figure}

\end{document}